\documentclass[11pt,a4paper]{article}
\usepackage{times,latexsym}
\usepackage{url}
\usepackage[T1]{fontenc}
\usepackage[utf8]{inputenc}
\DeclareUnicodeCharacter{0307}{}

\usepackage[acceptedWithA]{tacl2021v1}

\usepackage{graphicx}
\usepackage[export]{adjustbox}
\usepackage{microtype}
\usepackage{makecell}
\usepackage{booktabs}
\usepackage{multirow}
\usepackage{multicol}
\usepackage{colortbl}
\usepackage{amsmath} 
\usepackage{soul}
\usepackage{enumitem}

\usepackage{tikz}
\usetikzlibrary{shapes, arrows.meta, positioning}
\usetikzlibrary{calc}
\tikzstyle{stuff_nofill}=[rectangle,draw,font={A}]
\tikzstyle{stuff_fill}=[rectangle,draw,fill=gray!5,minimum size=1.4em]

\definecolor{pos}{HTML}{DAE8FC}
\definecolor{neg}{HTML}{F8CECC}

\DeclareRobustCommand{\hlpos}[1]{{\sethlcolor{pos}\hl{#1}}}
\DeclareRobustCommand{\hlneg}[1]{{\sethlcolor{neg}\hl{#1}}}

\title{Investigating Hallucinations in Pruned\\ Large Language Models for Abstractive Summarization}

\author{George Chrysostomou\Thanks{Equal contribution.}$^{*\clubsuit}$\Thanks{Work done independently of AstraZeneca.} \quad Zhixue Zhao$^{*\diamondsuit}$ \quad Miles Williams$^{*\diamondsuit}$ \quad Nikolaos Aletras$^{\diamondsuit}$ \\ $^{\diamondsuit}$University of Sheffield, UK \quad $^{\clubsuit}$AstraZeneca, UK \\ \texttt{\{zhixue.zhao, mwilliams15, n.aletras\}@sheffield.ac.uk}}

\begin{document}

\maketitle

\begin{abstract}

Despite the remarkable performance of generative large language models (LLMs) on abstractive summarization, they face two significant challenges: their considerable size and tendency to hallucinate. Hallucinations are concerning because they erode reliability and raise safety issues. Pruning is a technique that reduces model size by removing redundant weights, enabling more efficient sparse inference. Pruned models yield downstream task performance comparable to the original, making them ideal alternatives when operating on a limited budget. 
However, the effect that pruning has upon hallucinations in abstractive summarization with LLMs has yet to be explored.
In this paper, we provide an extensive empirical study across five summarization datasets, two state-of-the-art pruning methods, and five instruction-tuned LLMs. Surprisingly, we find that hallucinations are less prevalent from pruned LLMs than the original models. Our analysis suggests that pruned models tend to depend more on the source document for summary generation.  
This leads to a higher lexical overlap between the generated summary and the source document, which could be a reason for the reduction in hallucination risk.\footnote{\url{https://github.com/casszhao/PruneHall}}
\end{abstract}

\section{Introduction}

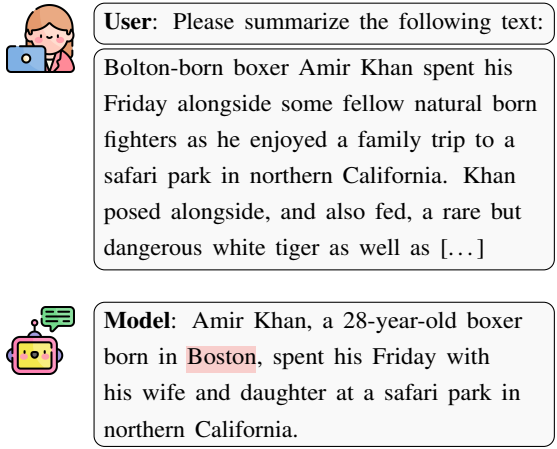
\begin{figure}[!t]
\centering
\begin{tikzpicture}
    \node at (-3,-0.20) {\includegraphics[width=0.9cm]{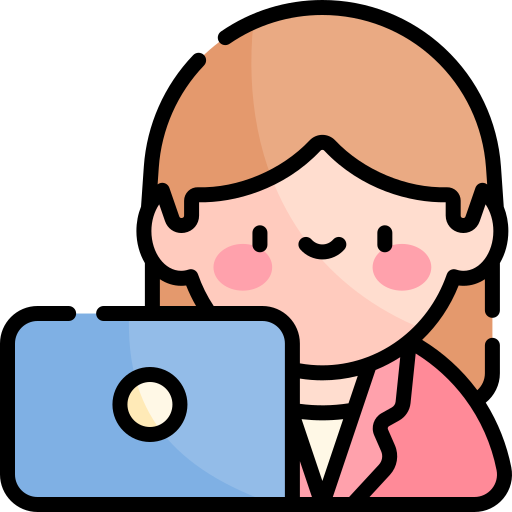}};
    
    \node[draw, rectangle, rounded corners=4pt, text width=\columnwidth-1.9cm, align=left, anchor=west] (user) at (-2.3,0) [stuff_fill] {\baselineskip=2pt \footnotesize \textbf{User}: Please summarize the following text:};

    \node[draw, rectangle, rounded corners=4pt, text width=\columnwidth-1.9cm, align=left, below=1.8cm of user, anchor=west] (model) at (-2.3,0) [stuff_fill] {\baselineskip=5pt \footnotesize Bolton-born boxer Amir Khan spent his Friday alongside some fellow natural born fighters as he enjoyed a family trip to a safari park in northern California. Khan posed alongside, and also fed, a rare but dangerous white tiger as well as [\textellipsis]};
    
    \node at (-3,-4.20) {\includegraphics[width=0.9cm]{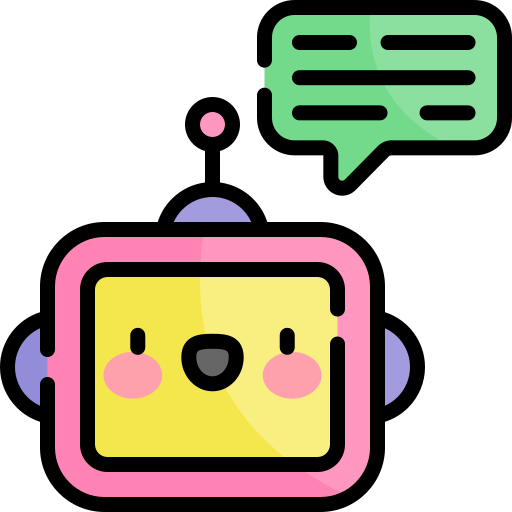}};
    
    \node[draw, rectangle, rounded corners=4pt, text width=\columnwidth-1.9cm, align=left, below=4.65cm of user, anchor=west] (model) at (-2.3,0) [stuff_fill] {\footnotesize \textbf{Model}: Amir Khan, a 28-year-old boxer born in \hlneg{Boston}, spent his Friday with his wife and daughter at a safari park in northern California.};
    
\end{tikzpicture}
\caption{An example of a hallucination (\hlneg{highlighted text}) in abstractive summarization.}
\label{fig:hallucination_example}
\end{figure}

Abstractive summarization is the task of distilling the key information from a document into a summary that may contain novel text not present in the original document \citep{cohn-lapata-2008-sentence, saggion-poibeau-2013-automatic, lin-ng-2019-abstractive}.
Generative large language models (LLMs) have demonstrated strong performance on abstractive summarization~\cite{ouyang2022training, touvron-etal-2023-llama-2, almazrouei-etal-2023-falcon,openai-2023-gpt4, zhang-etal-2024-benchmarking}. However, they face two significant challenges: their substantial size requires extensive computational resources for training and inference; and they tend to hallucinate, i.e. generate nonfactual contents not supported by the source document \citep{zhao-etal-2020-reducing, xu-etal-2023-understanding}. 
Figure \ref{fig:hallucination_example} shows an illustrative example of hallucinated content in a generated summary. 

On the one hand, hallucinations not only undermine the performance of models but also introduce critical safety risks, ultimately eroding the trust of end users \citep{milintsevich-agarwal-2023-calvados, tang-etal-2023-understanding, narayan-etal-2023-conditional,zhao2024reagent}. For example, LLM-generated summaries in the legal or health domain can contain inaccurate information that poses real-life harms \citep{zhao-etal-2022-impact,weidinger-etal-2022-taxonomy}. 

On the other hand, LLMs such as GPT-3.5 \citep{ouyang2022training}, GPT-4 \citep{openai-2023-gpt4}, and Llama-2 \citep{touvron-etal-2023-llama-2} demand substantial hardware resources. As an indication, GPT-3 (175B) requires at least five NVIDIA A100 GPUs with 80GB of memory each for half-precision inference \citep{frantar-alistarh-2023-sparsegpt}. This creates barriers for those without access to costly computational resources, ultimately hindering inclusivity in NLP \citep{schwartz-etal-2020-green, weidinger-etal-2022-taxonomy}. To tackle this issue, pruning techniques enable efficient sparse inference by removing redundant weights, while maintaining comparable performance \citep{sun-etal-2024-simple}. Pruned models therefore appear as attractive alternatives for abstractive summarization when computational resources are constrained. 

In abstractive summarization, model hallucinations are a thoroughly studied subject \citep{cao-etal-2020-factual, durmus-etal-2020-feqa, raunak-etal-2021-curious, narayan-etal-2023-conditional, laban-etal-2023-summedits}. Similarly, the effect of pruning on model performance in abstractive summarization benchmarks was also explored more recently \citep{dun2023sweeping, jaiswal-etal-2024-compressing}.
However, the relationship between pruning and hallucination risk has yet to be explored. Given the appeal of greater efficiency with comparable downstream performance it is important to establish how trustworthy summaries generated from pruned models are. Therefore, we seek to answer the following question: \emph{are hallucinations more or less prevalent in LLMs after pruning?}

To this end, we empirically investigate the risk of generating hallucinated content in pruned models across five LLMs, two state-of-the-art pruning methods, and five summarization datasets. Surprisingly, our results show that pruned models are less prevalent in hallucinations compared to the original LLM. To understand this phenomenon, we further investigate the impact of different sparsity levels on hallucination patterns. Our analysis shows that hallucination risk decreases as sparsity increases, regardless of the pruning methods tested. Furthermore, our results suggest that pruning encourages the model to rely more on the source document during generation, resulting in summaries that are lexically more similar to the source document.

\section{Related Work}

\subsection{Hallucinations in Summarization}
In abstractive summarization, a model is expected to generate a concise summary of the source document. However, prior work observed that models tend to generate hallucinatory content that is not based on or cannot be entailed from the source document \citep{vinyals-le-2015-neural, rohrbach-etal-2018-object, cao-etal-2018-faithful, maynez-etal-2020-faithfulness, raunak-etal-2021-curious, falke-etal-2019-ranking, maynez-etal-2020-faithfulness, zhao2022utilizing, chen-etal-2022-towards-improving}.
For example, \citet{falke-etal-2019-ranking} found that 25\% of the model generated summaries contain hallucinated content. On the other hand, automatic summary quality evaluation metrics such as ROUGE \citep{lin-2004-ROUGE} and BERTScore \citep{zhang-etal-2020-bertscore} do not correlate with the degree of hallucinations appearing in summaries \citep{zhou-etal-2021-detecting}. For instance, \citet{zhou-etal-2021-detecting} show that even if a summary contains a large amount of hallucinatory content, it can still achieve a high ROUGE score. This has opened up new research directions that develop approaches to detect and evaluate hallucinations \citep{zhou-etal-2021-detecting, durmus-etal-2020-feqa, guerreiro-etal-2023-optimal, ji-etal-2023-towards}, as well as mitigate them \citep{xiao-wang-2021-hallucination, choubey-etal-2023-cape, king-etal-2022-dont}.

\subsection{Measuring Hallucination Risk}

Evaluation metrics for measuring hallucination risk can be broadly categorized as: (a) entailment-based, (b) question-answering (QA), and (c) text-generation based. Entailment-based methods \citep{kryscinski-etal-2020-evaluating, laban-etal-2022-summac}  use pre-trained language models to compute the entailment score between the source and the generated summary. The higher the entailment score, the more consistent a summary is with respect to the source. QA methods decompose the task to a question answering problem \citep{wang-etal-2020-asking, deutsch-etal-2021-towards, durmus-etal-2020-feqa}. 
Finally, text-generation based methods use off-the-shelf models to quantify the risk of hallucinations \citep{yuan-metal-2021-bartscore, son-etal-2022-harim}. A representative approach is the Hallucination Risk Measurement (HaRiM\textsuperscript{+}) which uses the log-likelihoods from a reference-free decoder model to estimate hallucination risk in a summary at the token level \citep{son-etal-2022-harim}. More recently, \citet{laban-etal-2023-summedits} examined instruction-tuned LLMs as reasoners for factual assessments (i.e. assessors of hallucination prevalence) in abstractive text summarization. They demonstrated that many of these LLMs struggle to compete with previous entailment-based methods.

\subsection{Pruning Large Language Models}

Model compression is the task of reducing the memory footprint of a model \citep{ganesh-etal-2021-compressing}.
Pruning is a popular technique that removes redundant weights from the model \citep{lecun-etal-1989-optimal}. Weights may be removed individually (unstructured pruning), according to defined blocks (semi-structured pruning), or in relation to model components (structured pruning) \citep{blablock-etal-2020-what, mishra-etal-2021-accelerating, ma-etal-2023-llm}.

As the size of LLMs surpasses billions of parameters, pruning techniques that require re-training become impractical. Instead, post-training compression aims to reduce model size using only a small calibration dataset \citep{nagel-etal-2020-up, williams-aletras-2023-how}. In this setting, \citet{frantar-alistarh-2022-optimal} define the layer-wise compression problem, with the aim of creating a compressed version of a given layer that functions as closely as possible to the original. State-of-the-art post-training pruning techniques, such as SparseGPT \citep{frantar-alistarh-2023-sparsegpt} and Wanda \citep{sun-etal-2024-simple}, build upon this, offering layer-wise solutions. SparseGPT introduces an efficient approximation that relies upon an iterative weight update process using Hessian inverses, inspired by Optimal Brain Surgeon \citep{hassibi-etal-1993-optimal}. 
Wanda further improves upon efficiency by avoiding a weight update procedure, enabling pruning in a single forward pass.

In practice, the sparsity induced by pruning enables substantial improvements in inference performance across a variety of hardware. On a CPU, \citet{frantar-alistarh-2023-sparsegpt} demonstrate a 1.82$\times$ speedup with 50\% unstructured sparsity, using the DeepSparse engine \citep{neural-magic-2021-deepsparse}. Separately, they observe a 1.54-1.79$\times$ speedup for feed-forward layers on an NVIDIA Ampere GPU, using 2:4 semi-structured sparsity \citep{mishra-etal-2021-accelerating}. 

Recent pruning approaches (such as SparseGPT and Wanda) can be applied to decoder-only LLMs with minimal impact upon common-sense reasoning \citep{sun-etal-2024-simple} or summarization performance \citep{jaiswal-etal-2024-compressing}. Interestingly, related studies suggest that pruning can reduce social bias and toxicity \citep{xu-hu-2022-can} and improve resilience to `jailbreaking' attacks \citep{hasan-etal-2024-pruning}.
However, it remains unclear how pruning affects hallucination risk in LLMs.

\section{Methodology}
\label{sec:methodology}

\subsection{Models}

We experiment with the following publicly available LLMs: (1) the \textbf{Llama-2} \citep{touvron-etal-2023-llama-2} model family  (7B, 13B, and 70B); (2) \textbf{Mistral} 7B (v0.1) \citep{jiang2023mistral}; (3) \textbf{Falcon} 7B \citep{almazrouei-etal-2023-falcon}; and (4) the \textbf{OPT-IML} \citep{iyer-etal-2022-opt} model family (1.3B and 30B). 

We opt for decoder-only instruction-tuned models due to their efficacy in zero-shot abstractive summarization tasks \citep{tang-etal-2023-context, adams-etal-2023-sparse, laskar-etal-2023-building}.

\subsection{Pruning Methods}

We consider three different pruning methods: one standard baseline (layer-wise magnitude) and two state-of-the-art techniques (SparseGPT and Wanda). Formally, these pruning methods provide a saliency score $\mathbf{S}_{ij}$ for each element of the weight matrix $\mathbf{W}_{ij}$ in a given layer. The elements corresponding to the $k$ smallest saliency scores are the target weights to be pruned, where $k$ is determined by the sparsity ratio. The primary distinction between our selected pruning methods lies in their saliency score calculation metrics. In a post-training setting, pruning metrics can additionally incorporate layer activations, $\mathbf{X}$. The activations for each layer of the model are computed through performing a forward pass with the calibration data. We follow \citet{sun-etal-2024-simple} in using the same calibration data for each model, specifically 128 examples randomly sampled from C4 \citep{raffel2020exploring}.

\paragraph{Magnitude \normalfont{\citep{hagiwara-1994-simple, han-etal-2015-learning}}}

To offer a lower bound for the performance of pruned models, we employ layer-wise weight magnitude pruning. Here, the saliency score is simply the magnitude of each weight: 
$$\mathbf{S}_{ij} = |\mathbf{W}_{ij}|$$

\paragraph{SparseGPT \normalfont{\citep{frantar-alistarh-2023-sparsegpt}}} The SparseGPT algorithm is an iterative procedure that offers an efficient approximation to the exact layer reconstruction. The effective saliency criterion is 
$$\mathbf{S}_{ij} = \left[ |\mathbf{W}|^2/\text{diag} \left((\mathbf{XX}^T + \lambda \mathbf{I})^{-1} \right) \right]_{ij}$$
where $\lambda$ is a dampening parameter to enable inversion of the Hessian, $\mathbf{XX}^T + \lambda \mathbf{I}$.\footnote{We follow \citet{frantar-alistarh-2023-sparsegpt} in using $\lambda = 0.01$.}

\paragraph{Wanda \normalfont{\citep{sun-etal-2024-simple}}}

In contrast, Wanda avoids a computationally expensive weight update procedure, instead relying upon only the weight magnitudes and norm of the input activations: $$\mathbf{S}_{ij} = |\mathbf{W}_{ij}| \cdot ||\mathbf{X}||_2$$ This approximates SparseGPT when considering only diagonal elements of the Hessian for $\lambda = 0$.

\paragraph{Sparsity Level}

Following previous work \citep{frantar-alistarh-2023-sparsegpt, sun-etal-2024-simple}, we evaluate our pruning methods across both semi-structured and unstructured settings:

\begin{itemize}
    \item \textbf{2:4 semi-structured sparsity}: Two weights in every contiguous block of four must be zero, providing a total of 50\% sparsity. This sparsity pattern is required to enable hardware acceleration on GPUs \citep{mishra-etal-2021-accelerating}.
    \item \textbf{50\% unstructured sparsity}: To enable comparison, we use a sparsity level of 50\% for unstructured pruning, unless otherwise stated.
\end{itemize}

\begin{table}[!t]
\centering
\scriptsize
\begin{tabular}{lp{6.5cm}}
\toprule
\# & Prompt Template \\
\midrule
A & \textit{Summarize in a single short paragraph the context below:} \newline
\texttt{[document]} \newline
\textit{The summary is:} \texttt{[summary]} \\
\midrule
B & \textit{Summarize in a couple of sentences the document below:} \newline
\texttt{[document]} \newline
\textit{The summary is:} \texttt{[summary]} \\
\midrule
C & \textit{Give me a short summary of the below:} \newline
\texttt{[document]} \newline
\textit{The summary is:} \texttt{[summary]} \\
\bottomrule
\end{tabular}
\caption{Each prompt template consists of the task instructions (\textit{italic}) and the source \texttt{[document]}. The LLM then generates the \texttt{[summary]}.}
\label{tab:prompt_templates}
\end{table}

\noindent We do not explore pruning above 50\% sparsity as language modeling performance collapses shortly beyond this threshold \citep{frantar-alistarh-2023-sparsegpt, sun-etal-2024-simple}. Maintaining language modeling performance is essential for the generation of high-quality summaries, enabling comparison between the models and their pruned counterparts.

\subsection{Prompting}

LLMs are known to be sensitive to prompt design \citep{petroni-etal-2019-language,elazar-etal-2021-measuring,fierro-sogaard-2022-factual}. To mitigate the effect of prompt variability, we summarize each document using three distinct prompt templates (Table~\ref{tab:prompt_templates}). Each template instructs the model to summarize a given document in a slightly different manner, offering three summaries for each document. We then evaluate all three summaries by averaging the scores.

For each model family, we follow the prompt formatting used in the original work. In the case of Llama-2 and Mistral, this includes the use of \texttt{[INST]} and \texttt{[/INST]} tokens to delimit user instructions. For the Falcon and OPT-IML model families, which were not trained with a specific prompt format, we use the prompts as is (Table \ref{tab:prompt_templates}).

\subsection{Summarization Datasets}

We include the following summarization datasets: (1) \textbf{FactCC} \citep{kryscinski-etal-2020-evaluating}; (2) \textbf{Polytope} \citep{huang-etal-2020-achieved}; (3) \textbf{SummEval} \citep{fabbri-etal-2021-summeval}; (4) \textbf{Legal Contracts} \citep{manor-li-2019-plain}; and (5) \textbf{RCT} summaries \citep{wallace2021generating}. FactCC, Polytope and SummEval are all different subsets of the CNN/DailyMail news article dataset \citep{nallapati-etal-2016-abstractive}, covering a variety of topics. Legal Contracts consists of legal text snippets from the terms of service for various products and services. Finally, RCT combines the abstracts from randomized control trials with their corresponding human-written conclusions from systematic reviews, i.e. the conclusions are used as the target summary. For simplicity, we select instances in RCT where there is a one-to-one mapping between abstract and target summary.

We use the test set from each dataset and remove any duplicates if any exist. 
Table \ref{tab:dataset_statistics} provides detailed dataset statistics.

\subsection{Evaluation of Summarization Quality}

We evaluate the quality of generated summaries against the corresponding reference summary, using a subset of the ROUGE family of metrics \citep{lin-2004-ROUGE} and BERTScore \citep{zhang-etal-2020-bertscore}.\footnote{For FactCC, we use the extracted claim as the reference.} From ROUGE, we use two $n$-gram overlap metrics (ROUGE-1 and ROUGE-2) and the longest sequence overlap metric (ROUGE-L).

\begin{table}[t]
\centering
\scriptsize
    \begin{tabular}{lrrrrr}
    \toprule
    & & \multicolumn{2}{c}{Source} & \multicolumn{2}{c}{Reference} \\
    \cmidrule(lr){3-4}
    \cmidrule(lr){5-6}
    Dataset & \multicolumn{1}{c}{\#} & \multicolumn{1}{c}{Mean} & \multicolumn{1}{c}{Max} & \multicolumn{1}{c}{Mean} & \multicolumn{1}{c}{Max} \\
    \midrule
    FactCC & 311 & 634.2 & 1838 & 17.4 & 63 \\
    Polytope & 634 & 575.1 & 1781 & 64.6 & 128 \\
    SummEval & 100 & 407.8 & 589 & 65.1 & 101 \\
    Legal Contracts & 85 & 237.8 & 1106 & 21.6 & 61 \\
    RCT & 53 & 307.5 & 447 & 68.7 & 174 \\
    \bottomrule
    \end{tabular}
\caption{The number of source documents in each dataset (\#), and the mean and maximum length (in words) for the documents and reference summaries.}
\label{tab:dataset_statistics}
\end{table}

\subsection{Hallucination Risk Metrics}

To automatically evaluate the hallucination risk in the generated summaries, we use standard automatic metrics that compare directly the source document and the corresponding generated summary. 

\paragraph{HaRiM\textsuperscript{+} \normalfont{\citep{son-etal-2022-harim}}} HaRiM is based on the idea that over-reliance on decoder context during generation leads to hallucinations. Given a summary and a reference document, HaRiM\textsuperscript{+} first uses a pre-trained sequence-to-sequence model (S2S, an encoder-decoder model) to calculate the token probabilities in the summary given the reference document as input. 
A pre-trained decoder-only model is used as a secondary model (Aux) to compute summary token probabilities, i.e. no input document is provided to summarize. HaRiM\textsuperscript{+} therefore uses Aux token probabilities to regularize S2S token probabilities and detect hallucinations by:
$$\text{HaRiM} = \frac{1}{L} \sum^L_{i=0} (1 - p_{\text{S2S}}) (1 - (p_{\text{S2S}} - p_{\text{Aux}}))$$ where $L$ is the sequence length, $p_{\text{S2S}}$ the predicted probability of a token generated by the model given the source document, and $p_{\text{Aux}}$ the probability of the same generated token from the auxiliary model. 

HaRiM\textsuperscript{+} extends HaRiM through adding the S2S log-likelihood of tokens, and applying a scaling hyperparameter $\lambda_{H}$:\footnote{We follow \citet{son-etal-2022-harim} in using $\lambda_{H} = 7$.} 
$$\text{HaRiM}^{\text{+}} = \frac{1}{L}\sum_i^L \text{log}(p(y_i \mid y_{<i}; X)) - \lambda_{H} \text{HaRiM}$$

\noindent Intuitively, a higher HaRiM\textsuperscript{+} score indicates that the summary is more likely to be faithful to the source document, i.e. less likely to contain hallucinations. \citet{son-etal-2022-harim} also showed that the first sequence-to-sequence model can also act as a secondary model, with equivalent performance.

\paragraph{SummaC \normalfont{\citep{laban-etal-2022-summac}}} This metric uses an off-the-shelf entailment model to assess the consistency between a source document and a generated summary.  
First, the document and summary are split into sentences, with the document sentences ($N$) being the hypothesis and the generated summary sentences ($K$) being the premise. The second step is to create an $K \times N$ matrix of entailment scores from the pre-trained model. A generated sentence with a low entailment score to any of the document sentences is a potential hallucination.
\par
\noindent\textbf{SummaC\textsubscript{ZS}} obtains the row-wise maximum entailment score, which leads to a vector $E$ of size $K$.
\textbf{SummaC\textsubscript{Conv}} obtains vector $E$ by using a convolutional model over each row $K$, to obtain a single score. In both metrics, each element in $E$ can be interpreted as the consistency score for each sentence in the summary. $E$ is averaged to obtain a single summary consistency score.

\paragraph{Hallucination Risk Ratio (HRR)}
\label{sec:hrr}

To compare the hallucination risk of pruned models relative to the original, we compute a ratio using any one of the hallucination risk metrics: $$\text{HRR} = \frac{\text{Hallucination Risk}_{\text{Original}}}{\text{Hallucination Risk}_{\text{Pruned}}}$$
\noindent A lower HRR indicates that the pruned model has a lower hallucination risk than the original. This contrasts the hallucination risk metrics, where a higher score indicates a lower risk for a given model.

\begin{table}[!t]
\centering
\renewcommand\thetable{3}
\resizebox{\columnwidth}{!}{
\begin{tabular}{lrrrrrrr}
\toprule
& & \multicolumn{2}{c}{Magnitude} & \multicolumn{2}{c}{SparseGPT} & \multicolumn{2}{c}{Wanda} \\
\cmidrule(lr){3-4}
\cmidrule(lr){5-6}
\cmidrule(lr){7-8}
Model &\multicolumn{1}{c}{-} & \multicolumn{1}{c}{2:4} & \multicolumn{1}{c}{50\%} & \multicolumn{1}{c}{2:4} & \multicolumn{1}{c}{50\%} & \multicolumn{1}{c}{2:4} & \multicolumn{1}{c}{50\%} \\
\midrule
Falcon 7B & 19.93 & 303.22 & 482.11 & 52.11 & 37.10  & 85.68 & 38.93 \\
Llama-2 7B & 6.49  & 78.29 & 19.07 & 10.79 & 7.94 & 12.46 & 7.93\\
Llama-2 13B & 5.71 & 10.73 & 7.98 & 8.68 & 6.80 &  9.58 & 6.94 \\
Llama-2 70B & 4.30 & 6.89 & 5.61 & 6.51 & 5.18 & 6.45 & 5.23 \\
Mistral 7B &  6.32 & 9.55 & 7.96 & 9.21 & 7.18  &  9.85 & 7.26 \\
OPT-IML 1.3B & 14.68 & 166.09 & 1391.46 & 24.92 & 18.03 & 25.11 & 17.94 \\ 
OPT-IML 30B & 10.56 & 246.42 & 57.88 & 11.61 & 10.74& 12.44 & 10.74 \\
\bottomrule
\end{tabular}
}
\caption{Perplexity ($\downarrow$) of original and pruned models on the held-out set of WikiText.}
\label{tab:performance_of_compressed_models}
\end{table}

\renewcommand{\arraystretch}{1.30}
\begin{table*}[!t]
    \centering
    \renewcommand\thetable{4}
    \resizebox{\linewidth}{!}{%
    \begin{tabular}{lc||cc|cc|cc|cc|cc}
\toprule
& & \multicolumn{2}{c|}{Llama-2 7B} & \multicolumn{2}{c|}{Llama-2 13B} & \multicolumn{2}{c|}{Llama-2 70B} & \multicolumn{2}{c|}{Mistral 7B} & \multicolumn{2}{c}{OPT-IML 30B} \\\
Dataset & Method & ROUGE-1/2/L & BS & ROUGE-1/2/L & BS & ROUGE-1/2/L & BS  & ROUGE-1/2/L & BS & ROUGE-1/2/L & BS \\
\midrule \midrule

\multirow{3}{*}{FactCC} & - & 13.99 / 6.41 / 11.51 & 84.60 & 15.14 / 6.39 / 12.30 & 84.39 & 15.04 / 6.29 / 12.11 & 84.75 & 14.83 / 8.21 / 12.70 & 84.78 & 23.51 / 12.68 / 20.48 & 85.71 \\

& SpGPT &\underline{12.46} / 6.07 / \underline{10.55} & \underline{84.15} & 15.34 / 6.62 / 12.75 & \textbf{84.76} & 14.78 / 6.80 / 12.29 & 84.68 & 14.43 / 8.52 / 12.62 & 84.45 & \underline{18.52} / 12.05 / \underline{16.89} & \underline{85.04} \\
& Wanda & \underline{11.04} / 5.94 / \underline{9.53} & \underline{80.57} & 15.64 / 7.32 / 13.09 & \textbf{84.78} & 15.09 / 6.88 / 12.47 & 84.72 & 13.67 / 8.30 / 12.02 & \underline{84.34} & \underline{17.91} / 11.68 / \underline{16.38} & \underline{83.94} \\
\hline

\multirow{3}{*}{Polytope} & - & 38.92 / 18.19 / 25.86 & 85.41 & 38.63 / 17.51 / 25.34 & 84.91 & 39.28 / 17.48 / 25.78 & 85.48 & 40.27 / 22.69 / 28.65 & 85.63 & 33.06 / 22.81 / 27.74 & 86.54 \\

&SpGPT & \underline{33.98} / 18.14 / \underline{24.45} & \underline{84.88} & \underline{35.99} / 16.74 / 25.01 & 85.01 & 38.16 / 18.51 / 25.89 & 85.31 & 39.07 / \textbf{24.21} / 29.54 & 85.58 & 33.39 / \textbf{26.32} / 29.02 & \textbf{87.01} \\
& Wanda & \underline{30.88} / \underline{15.39} / \underline{21.77} & \underline{83.09} & \underline{37.33} / \textbf{19.29} / \textbf{26.68} & \textbf{85.23} & 38.74 / \textbf{18.80} / 26.58 & 85.42 & \underline{37.08} / 23.78 / 28.76 & \underline{85.34} & \underline{30.14} / 22.72 / \underline{25.85} & \underline{86.03} \\ \hline

\multirow{3}{*}{SummEval}  & - & 40.39 / 18.73 / 26.61 & 85.42 & 40.36 / 18.00 / 25.88 & 84.78 & 41.52 / 18.78 / 26.82 & 85.58 & 43.94 / 26.34 / 32.04 & 86.05 & 51.93 / 36.55 / 41.38 & 86.94 \\
& SpGPT & 38.77 / \textbf{23.04} / 27.81 & 85.36 & 40.55 / 18.42 / 27.15 & \textbf{85.33} & 41.58 / 19.69 / 27.65 & 85.61 & 43.77 / 28.00 / 33.33 & 86.03 & 50.00 / 37.16 / 41.64 & 86.73 \\
& Wanda & 37.78 / \textbf{23.95} / 28.82 & 85.12 & \textbf{44.31} / \textbf{23.51} / \textbf{31.58} & \textbf{86.03} & 41.57 / 19.44 / 27.67 & 85.57 & 45.11 / 29.95 / 34.84 & 86.22 & \underline{44.48} / 33.57 / 36.90 & \underline{86.12} \\ \hline

\multirow{3}{*}{\makecell[l]{Legal \\ Contracts}} & - & 18.75 / 6.20 / 13.93 & 84.73 & 21.12 / 6.90 / 15.41 & 84.75 & 21.66 / 7.07 / 16.19 & 85.60 & 17.52 / 6.21 / 13.70 & 84.78 & 22.96 / 7.45 / 18.30 & 84.90 \\
& SpGPT & 16.84 / 5.98 / 12.80 & 84.17 & 18.99 / 6.11 / 14.41 & 84.90 & 21.74 / 7.42 / 16.73 & 85.33 & 18.56 / 6.90 / 14.51 & 84.76 & 21.18 / 7.22 / 17.15 & 84.49 \\
& Wanda & \underline{14.22} / 4.94 / 11.14 & \underline{81.52} & 18.80 / 6.37 / 14.53 & 84.41 & 22.13 / 7.51 / 16.72 & 85.55 & 18.14 / 6.37 / 13.83 & 84.79 & 19.10 / 6.79 / 15.36 & \underline{81.86} \\ \hline

\multirow{3}{*}{RCT} & - & 45.29 / 26.89 / 33.50 & 86.97 & 39.87 / 22.01 / 28.56 & 86.43 & 37.79 / 20.98 / 28.05 & 86.25 & 53.66 / 40.66 / 46.16 & 88.46 & 24.62 / 18.20 / 21.33 & 83.12 \\
& SpGPT &\textbf{50.57} / \textbf{37.40} / \textbf{43.12} & \textbf{87.89} & \underline{37.81} / 22.40 / 29.37 & 86.26 & \textbf{40.19} / \textbf{25.35} / \textbf{31.97} & \textbf{86.57} & \textbf{56.93} / \textbf{47.79} / \textbf{52.45} & \textbf{89.17} & 25.22 / \textbf{21.50} / 23.61 & \underline{77.39} \\
& Wanda & \underline{38.79} / 28.59 / 33.12 & \underline{86.06} & \underline{36.90} / 23.07 / 28.82 & \underline{86.11} & \textbf{39.61} / \textbf{24.79} / \textbf{31.60} & 86.49 & \textbf{59.29} / \textbf{50.02} / \textbf{54.83} & \textbf{89.40} & \textbf{31.59} / \textbf{28.84} / \textbf{30.49} & \underline{70.64} \\

\bottomrule
\end{tabular}
}
    \caption{ROUGE-1/2/L ($\uparrow$) and BERTScore (BS; $\uparrow$) for the original models (-) and their pruned counterparts (SparseGPT and Wanda). Values in \textbf{bold} indicate that the pruned model scores significantly higher than the original while  \underline{underlined} values denote a significantly lower score (paired t-test; $p < 0.05$).}
    \label{tab:generation_performance_main}
\end{table*}

\subsection{Human Evaluation}\label{sec:human_evaluation_method}

We also conduct a human evaluation task to compare the hallucination prevalence between the original and pruned models. For this purpose, we randomly sample 100 distinct source documents from FactCC, Polytope and SummEval. We selected these datasets because they consist of news articles, making them suitable for human evaluation without requiring extensive domain expertise. 
We recruited three participants who are native speakers or proficiently fluent in English. Following \citet{lango-dusek-2023-critic}, we ask them to answer the following questions for comparing the summaries generated by the original and pruned models: 

\begin{enumerate}[label=Q\arabic*.]
\item \textbf{Hallucinations}: 
Which summary contains more hallucinations (i.e. content that is not supported by the source document)?
\item \textbf{Omission}: Which summary is missing more crucial information from the document?
\item \textbf{Repetition}: Which summary contains more repetitive information?
\item \textbf{Alignment}: Which summary is more semantically aligned with the source document?
\end{enumerate}

Identifying hallucinations in text is challenging and requires careful reading and attention to nuanced facts~\citep{laban-etal-2023-summedits}. Therefore, we first perform a calibration run on a held-out set of ten documents and their generated summaries. Two of the participants are then presented with the set of 100 original documents, alongside two generated summaries: one from a pruned model and the other from the original model. The order of the documents is shuffled and information about which model generated the summary is not disclosed to the participants. Similar to \citet{xu-etal-2023-understanding}, we use the third participant as an adjudicator for disagreements. The inter-annotator agreement is computed using Cohen’s kappa IAA ($\kappa$), as the average between the two participants and the adjudicator. 

\subsection{Implementation Details}

We use the model implementation and weights available from Hugging Face \citep{wolf-etal-2020-transformers}. We perform experiments using either one or two NVIDIA A100 (SXM 80GB) GPUs. For the pruning methods, we use the hyperparameters from \citet{frantar-alistarh-2023-sparsegpt} and \citet{sun-etal-2024-simple}.

For summary generation we use greedy decoding (i.e. sampling the token with the highest probability) for better reproducibility.
We continue to sample tokens until we reach either (a) the end of sequence token, or (b) the maximum sequence length of the model.

\section{Results}

\subsection{Language Modeling}

We first compare language modeling performance between the original and pruned models. Following \citet{frantar-alistarh-2023-sparsegpt} and \citet{sun-etal-2024-simple}, we compute perplexity on the WikiText test set \citep{merity-etal-2017-pointer}, shown in Table \ref{tab:performance_of_compressed_models}.

Overall, pruned models consistently generate text with higher perplexity than their original counterparts. Unsurprisingly, magnitude pruning routinely produces the highest perplexity. In many cases, the increase over the original model (denoted by `-') is substantial. For example, we observe more than a twentyfold increase for OPT-IML 30B, from 10.56 to 246.42. In contrast, SparseGPT and Wanda achieve perplexity close to the original for the majority of the models. Surprisingly, Falcon 7B records higher perplexity across all pruning methods, e.g. 85.68 when applying Wanda from 19.93 without pruning.

Due to the substantial degradation in language modeling performance, we omit magnitude pruning from further analysis. For the same reason, we also exclude the Falcon 7B and OPT-IML 1.3B models.

\begin{table*}[!t]
    \centering
    \resizebox{\linewidth}{!}{
\begin{tabular}{ll||cccc|cccc|cccc|cccc|cccc}
\toprule
& & \multicolumn{4}{c|}{Llama-2 7B}  & \multicolumn{4}{c|}{Llama-2 13B} & \multicolumn{4}{c|}{Llama-2 70B} & \multicolumn{4}{c|}{Mistral 7B}& \multicolumn{4}{c}{OPT-IML 30B} \\
& & \multicolumn{2}{c}{SparseGPT} & \multicolumn{2}{c|}{Wanda} &  \multicolumn{2}{c}{SparseGPT} & \multicolumn{2}{c|}{Wanda}&  \multicolumn{2}{c}{SparseGPT} & \multicolumn{2}{c|}{Wanda} &  \multicolumn{2}{c}{SparseGPT} & \multicolumn{2}{c|}{Wanda} &  \multicolumn{2}{c}{SparseGPT} & \multicolumn{2}{c}{Wanda} \\ 
Dataset & Metric & 2:4 & 50\% & 2:4 & 50\% & 2:4 & 50\% & 2:4 & 50\% & 2:4 & 50\% & 2:4 & 50\% & 2:4 & 50\% & 2:4 & 50\% & 2:4 & 50\% & 2:4 & 50\% \\ \midrule \midrule

\multirow{3}{*}{FactCC}  & HaRiM\textsuperscript{+}& \cellcolor{teal!5} \textbf{0.98} & \cellcolor{teal!5} \textbf{0.95} & \cellcolor{teal!6} \textbf{0.94} & \cellcolor{teal!5} \textbf{0.95} & \cellcolor{teal!30} \textbf{0.77} & \cellcolor{teal!5} \textbf{0.95} & \cellcolor{teal!43} \textbf{0.69} & \cellcolor{teal!9} \textbf{0.91} & \cellcolor{teal!8} \textbf{0.93} & \cellcolor{teal!5} \textbf{0.96} & \cellcolor{teal!8} \textbf{0.93} & \cellcolor{teal!5} \textbf{0.96} & \cellcolor{teal!8} \textbf{0.93} & \cellcolor{teal!6} \textbf{0.94} & \cellcolor{teal!9} \textbf{0.91} & \cellcolor{teal!6} \textbf{0.94} & \cellcolor{teal!21} \textbf{0.83} & \cellcolor{teal!15} \textbf{0.87} & \cellcolor{teal!15} \textbf{0.87} & \cellcolor{teal!16} \textbf{0.85} \\
 & SummaC\textsubscript{conv} & \cellcolor{teal!55} \textbf{0.64} & \cellcolor{teal!22} \textbf{0.82} & \cellcolor{teal!55} \textbf{0.56} & \cellcolor{teal!23} \textbf{0.81} & \cellcolor{teal!32} \textbf{0.76} & \cellcolor{teal!19} \textbf{0.83} & \cellcolor{teal!55} \textbf{0.64} & \cellcolor{teal!18} \textbf{0.84} & \cellcolor{teal!32} \textbf{0.76} & \cellcolor{teal!8} \textbf{0.92} & \cellcolor{teal!30} \textbf{0.77} & \cellcolor{teal!10} \textbf{0.90} & \cellcolor{teal!25} \textbf{0.79} & \cellcolor{teal!14} \textbf{0.88} & \cellcolor{teal!34} \textbf{0.74} & \cellcolor{teal!16} \textbf{0.86} & \cellcolor{teal!25} \textbf{0.80} & \cellcolor{teal!16} \textbf{0.86} & \cellcolor{teal!18} \textbf{0.84} & \cellcolor{teal!21} \textbf{0.83} \\
 & SummaC\textsubscript{zs} & \cellcolor{teal!55} \textbf{0.47} & \cellcolor{teal!52} \textbf{0.65} & \cellcolor{teal!55} \textbf{0.39} & \cellcolor{teal!55} \textbf{0.65} & \cellcolor{teal!55} \textbf{0.50} & \cellcolor{teal!55} \textbf{0.61} & \cellcolor{teal!55} \textbf{0.41} & \cellcolor{teal!55} \textbf{0.61} & \cellcolor{teal!55} \textbf{0.63} & \cellcolor{teal!16} \textbf{0.86} & \cellcolor{teal!55} \textbf{0.63} & \cellcolor{teal!21} \textbf{0.83} & \cellcolor{teal!32} \textbf{0.76} & \cellcolor{teal!16} \textbf{0.85} & \cellcolor{teal!47} \textbf{0.68} & \cellcolor{teal!22} \textbf{0.82} & \cellcolor{teal!25} \textbf{0.80} & \cellcolor{teal!15} \textbf{0.87} & \cellcolor{teal!17} \textbf{0.85} & \cellcolor{teal!19} \textbf{0.83} \\ \hline

\multirow{3}{*}{Polytope}  & HaRiM\textsuperscript{+}& \cellcolor{teal!5} \textbf{0.97} & \cellcolor{teal!5} \textbf{0.97} & \cellcolor{teal!5} \textbf{0.97} & \cellcolor{teal!5} \textbf{0.97} & \cellcolor{teal!28} \textbf{0.78} & \cellcolor{teal!7} \textbf{0.93} & \cellcolor{teal!40} \textbf{0.71} & \cellcolor{teal!17} \textbf{0.85} & \cellcolor{teal!6} \textbf{0.94} & \cellcolor{teal!5}0.96 & \cellcolor{teal!5} \textbf{0.95} & 1.00 & \cellcolor{teal!5} \textbf{0.95} & \cellcolor{teal!5} \textbf{0.95} & \cellcolor{teal!6} \textbf{0.94} & \cellcolor{teal!5} \textbf{0.96} & \cellcolor{teal!15} \textbf{0.87} & \cellcolor{teal!7} \textbf{0.93} & \cellcolor{teal!8} \textbf{0.92} & \cellcolor{teal!12} \textbf{0.88} \\
 & SummaC\textsubscript{conv} & \cellcolor{teal!49} \textbf{0.67} & \cellcolor{teal!21} \textbf{0.83} & \cellcolor{teal!43} \textbf{0.69} & \cellcolor{teal!21} \textbf{0.83} & \cellcolor{teal!42} \textbf{0.70} & \cellcolor{teal!28} \textbf{0.78} & \cellcolor{teal!55} \textbf{0.65} & \cellcolor{teal!25} \textbf{0.79} & \cellcolor{teal!30} \textbf{0.77} & \cellcolor{teal!7} \textbf{0.93} & \cellcolor{teal!27} \textbf{0.78} & \cellcolor{teal!8} \textbf{0.92} & \cellcolor{teal!27} \textbf{0.78} & \cellcolor{teal!22} \textbf{0.82} & \cellcolor{teal!31} \textbf{0.76} & \cellcolor{teal!18} \textbf{0.84} & \cellcolor{teal!16} \textbf{0.86} & \cellcolor{teal!5} \textbf{0.95} & \cellcolor{teal!9} \textbf{0.91} & \cellcolor{teal!8} \textbf{0.92} \\
 & SummaC\textsubscript{zs} & \cellcolor{teal!55} \textbf{0.64} & \cellcolor{teal!17} \textbf{0.85} & \cellcolor{teal!55} \textbf{0.64} & \cellcolor{teal!33} \textbf{0.75} & \cellcolor{teal!55} \textbf{0.58} & \cellcolor{teal!44} \textbf{0.69} & \cellcolor{teal!55} \textbf{0.56} & \cellcolor{teal!44} \textbf{0.69} & \cellcolor{teal!33} \textbf{0.75} & \cellcolor{teal!14} \textbf{0.88} & \cellcolor{teal!34} \textbf{0.74} & \cellcolor{teal!19} \textbf{0.83} & \cellcolor{teal!32} \textbf{0.76} & \cellcolor{teal!24} \textbf{0.81} & \cellcolor{teal!33} \textbf{0.75} & \cellcolor{teal!18} \textbf{0.84} & \cellcolor{teal!14} \textbf{0.88} & \cellcolor{teal!5} \textbf{0.95} & \cellcolor{teal!8} \textbf{0.92} & \cellcolor{teal!8} \textbf{0.93} \\ \hline

 \multirow{3}{*}{SummEval} & HaRiM\textsuperscript{+}& \cellcolor{teal!12} \textbf{0.88} & \cellcolor{teal!8} \textbf{0.93} & \cellcolor{teal!24} \textbf{0.81} & \cellcolor{teal!8} \textbf{0.93} & \cellcolor{teal!25} \textbf{0.80} & \cellcolor{teal!5} \textbf{0.97} & \cellcolor{teal!43} \textbf{0.69} & \cellcolor{teal!5} \textbf{0.96} & \cellcolor{teal!5} \textbf{0.95} & \cellcolor{teal!5}0.98 & \cellcolor{teal!5} \textbf{0.95} & \cellcolor{teal!5}0.98 & \cellcolor{teal!7} \textbf{0.93} & \cellcolor{teal!6} \textbf{0.94} & \cellcolor{teal!8} \textbf{0.92} & \cellcolor{teal!5} \textbf{0.95} & \cellcolor{teal!9} \textbf{0.91} & \cellcolor{teal!8} \textbf{0.92} & \cellcolor{teal!10} \textbf{0.90} & \cellcolor{teal!12} \textbf{0.89} \\
 & SummaC\textsubscript{conv} & \cellcolor{teal!55} \textbf{0.55} & \cellcolor{teal!24} \textbf{0.81} & \cellcolor{teal!55} \textbf{0.46} & \cellcolor{teal!31} \textbf{0.76} & \cellcolor{teal!50} \textbf{0.67} & \cellcolor{teal!24} \textbf{0.81} & \cellcolor{teal!55} \textbf{0.59} & \cellcolor{teal!24} \textbf{0.81} & \cellcolor{teal!28} \textbf{0.78} & \cellcolor{teal!5}0.96 & \cellcolor{teal!25} \textbf{0.79} & \cellcolor{teal!7}0.93 & \cellcolor{teal!25} \textbf{0.79} & \cellcolor{teal!17} \textbf{0.85} & \cellcolor{teal!30} \textbf{0.77} & \cellcolor{teal!15} \textbf{0.87} & \cellcolor{teal!16} \textbf{0.86} & \cellcolor{teal!12} \textbf{0.88} & \cellcolor{teal!19} \textbf{0.83} & \cellcolor{teal!17} \textbf{0.85} \\
 & SummaC\textsubscript{zs} & \cellcolor{teal!55} \textbf{0.49} & \cellcolor{teal!33} \textbf{0.75} & \cellcolor{teal!55} \textbf{0.4} & \cellcolor{teal!47} \textbf{0.68} & \cellcolor{teal!55} \textbf{0.56} & \cellcolor{teal!41} \textbf{0.71} & \cellcolor{teal!55} \textbf{0.49} & \cellcolor{teal!50} \textbf{0.66} & \cellcolor{teal!41} \textbf{0.70} & \cellcolor{teal!8} \textbf{0.92} & \cellcolor{teal!41} \textbf{0.70} & \cellcolor{teal!12} \textbf{0.88} & \cellcolor{teal!26} \textbf{0.79} & \cellcolor{teal!18} \textbf{0.84} & \cellcolor{teal!32} \textbf{0.76} & \cellcolor{teal!14} \textbf{0.88} & \cellcolor{teal!16} \textbf{0.86} & \cellcolor{teal!12} \textbf{0.89} & \cellcolor{teal!17} \textbf{0.85} & \cellcolor{teal!16} \textbf{0.86} \\ \hline

\multirow{3}{*}{\makecell[l]{Legal \\ Contracts}} & HaRiM\textsuperscript{+}& \cellcolor{teal!5} \textbf{0.99} & \cellcolor{teal!16} \textbf{0.85} & \cellcolor{teal!10} \textbf{0.90} & \cellcolor{teal!16} \textbf{0.85} & \cellcolor{teal!19} \textbf{0.83} & \cellcolor{teal!12} \textbf{0.88} & \cellcolor{teal!32} \textbf{0.76} & \cellcolor{teal!12} \textbf{0.88} & \cellcolor{teal!15} \textbf{0.87} & \cellcolor{teal!8} \textbf{0.92} & \cellcolor{teal!12} \textbf{0.89} & \cellcolor{teal!5} \textbf{0.95} & \cellcolor{teal!16} \textbf{0.85} & \cellcolor{teal!6} \textbf{0.94} & \cellcolor{teal!12} \textbf{0.89} & \cellcolor{teal!8} \textbf{0.93} & \cellcolor{teal!16} \textbf{0.85} & \cellcolor{teal!12} \textbf{0.89} & \cellcolor{teal!24} \textbf{0.81} & \cellcolor{teal!19} \textbf{0.83} \\
 & SummaC\textsubscript{conv} & \cellcolor{teal!5}0.98 & \cellcolor{teal!17}0.85 & \cellcolor{teal!8}0.93 & \cellcolor{teal!6}0.94 & \cellcolor{teal!22} \textbf{0.82} & \cellcolor{teal!24} \textbf{0.81} & \cellcolor{teal!32} \textbf{0.76} & \cellcolor{teal!23} \textbf{0.81} & \cellcolor{teal!26} \textbf{0.79} & \cellcolor{teal!14} \textbf{0.88} & \cellcolor{teal!21} \textbf{0.83} & \cellcolor{teal!9} \textbf{0.91} & \cellcolor{teal!21} \textbf{0.83} & \cellcolor{teal!8} \textbf{0.92} & \cellcolor{teal!8} \textbf{0.92} & \cellcolor{teal!12} \textbf{0.89} & \cellcolor{teal!16} \textbf{0.85} & \cellcolor{teal!12} \textbf{0.88} & \cellcolor{teal!24} \textbf{0.81} & \cellcolor{teal!16} \textbf{0.86} \\
 & SummaC\textsubscript{zs} & \cellcolor{red!15}1.01 & \cellcolor{teal!16} \textbf{0.86} & \cellcolor{teal!5}0.96 & \cellcolor{teal!10} \textbf{0.90} & \cellcolor{teal!8} \textbf{0.93} & \cellcolor{teal!16} \textbf{0.86} & \cellcolor{teal!14} \textbf{0.88} & \cellcolor{teal!12} \textbf{0.88} & \cellcolor{teal!17} \textbf{0.85} & \cellcolor{teal!8}0.93 & \cellcolor{teal!14} \textbf{0.88} & \cellcolor{teal!5}0.95 & \cellcolor{teal!12} \textbf{0.88} & \cellcolor{teal!8} \textbf{0.92} & \cellcolor{teal!7} \textbf{0.93} & \cellcolor{teal!8} \textbf{0.92} & \cellcolor{teal!7} \textbf{0.93} & \cellcolor{teal!5}0.96 & \cellcolor{teal!6} \textbf{0.94} & 1.00 \\ \hline

 \multirow{3}{*}{RCT} & HaRiM\textsuperscript{+}& \cellcolor{teal!8} \textbf{0.92} & \cellcolor{teal!5} \textbf{0.96} & \cellcolor{teal!15} \textbf{0.87} & \cellcolor{teal!8} \textbf{0.92} & \cellcolor{teal!16} \textbf{0.86} & \cellcolor{teal!5} \textbf{0.99} & \cellcolor{teal!25} \textbf{0.80} & \cellcolor{teal!5} \textbf{0.97} & \cellcolor{teal!8} \textbf{0.93} & \cellcolor{teal!5} \textbf{0.96} & \cellcolor{teal!8} \textbf{0.93} & \cellcolor{teal!5} \textbf{0.97} & \cellcolor{teal!7} \textbf{0.93} & \cellcolor{teal!5} \textbf{0.96} & \cellcolor{teal!7} \textbf{0.93} & \cellcolor{teal!5} \textbf{0.95} & \cellcolor{teal!16} \textbf{0.85} & \cellcolor{teal!12} \textbf{0.88} & \cellcolor{teal!19} \textbf{0.83} & \cellcolor{teal!15} \textbf{0.87} \\
 & SummaC\textsubscript{conv} & \cellcolor{teal!44} \textbf{0.69} & \cellcolor{teal!16} \textbf{0.86} & \cellcolor{teal!41} \textbf{0.70} & \cellcolor{teal!12} \textbf{0.88} & \cellcolor{teal!27} \textbf{0.78} & \cellcolor{teal!12} \textbf{0.89} & \cellcolor{teal!26} \textbf{0.79} & \cellcolor{teal!12} \textbf{0.88} & \cellcolor{teal!22} \textbf{0.82} & \cellcolor{teal!8} \textbf{0.92} & \cellcolor{teal!22} \textbf{0.82} & \cellcolor{teal!7} \textbf{0.93} & \cellcolor{teal!22} \textbf{0.82} & \cellcolor{teal!12} \textbf{0.88} & \cellcolor{teal!24} \textbf{0.81} & \cellcolor{teal!15} \textbf{0.87} & \cellcolor{teal!19} \textbf{0.83} & \cellcolor{teal!12} \textbf{0.88} & \cellcolor{teal!26} \textbf{0.79} & \cellcolor{teal!14} \textbf{0.88} \\
 & SummaC\textsubscript{zs} & \cellcolor{teal!40} \textbf{0.71} & \cellcolor{teal!19} \textbf{0.83} & \cellcolor{teal!41} \textbf{0.71} & \cellcolor{teal!22} \textbf{0.82} & \cellcolor{teal!43} \textbf{0.69} & \cellcolor{teal!23} \textbf{0.81} & \cellcolor{teal!42} \textbf{0.70} & \cellcolor{teal!22} \textbf{0.82} & \cellcolor{teal!25} \textbf{0.79} & \cellcolor{teal!10} \textbf{0.90} & \cellcolor{teal!26} \textbf{0.79} & \cellcolor{teal!10} \textbf{0.90} & \cellcolor{teal!18} \textbf{0.84} & \cellcolor{teal!12} \textbf{0.89} & \cellcolor{teal!22} \textbf{0.82} & \cellcolor{teal!12} \textbf{0.89} & \cellcolor{teal!30} \textbf{0.77} & \cellcolor{teal!25} \textbf{0.80} & \cellcolor{teal!30} \textbf{0.77} & \cellcolor{teal!19} \textbf{0.83} \\ \hline\hline

 \multirow{3}{*}{Average} & HaRiM\textsuperscript{+}& \cellcolor{teal!5}0.95 & \cellcolor{teal!6}0.93 & \cellcolor{teal!9}0.90 & \cellcolor{teal!7}0.92 & \cellcolor{teal!18}0.81 & \cellcolor{teal!5}0.95 & \cellcolor{teal!27}0.73 & \cellcolor{teal!8}0.91 & \cellcolor{teal!7}0.92 & \cellcolor{teal!4}0.96 & \cellcolor{teal!6}0.93 & \cellcolor{teal!3}0.97 & \cellcolor{teal!7}0.92 & \cellcolor{teal!5}0.95 & \cellcolor{teal!7}0.92 & \cellcolor{teal!5}0.95 & \cellcolor{teal!14}0.87 & \cellcolor{teal!9}0.90 & \cellcolor{teal!13}0.87 & \cellcolor{teal!13}0.87 \\
& SummaC\textsubscript{conv} & \cellcolor{teal!30}0.70 & \cellcolor{teal!17}0.83 & \cellcolor{teal!32}0.67 & \cellcolor{teal!15}0.85 & \cellcolor{teal!26}0.74 & \cellcolor{teal!18}0.82 & \cellcolor{teal!31}0.68 & \cellcolor{teal!17}0.83 & \cellcolor{teal!21}0.78 & \cellcolor{teal!7}0.92 & \cellcolor{teal!19}0.80 & \cellcolor{teal!7}0.92 & \cellcolor{teal!19}0.80 & \cellcolor{teal!13}0.87 & \cellcolor{teal!19}0.80 & \cellcolor{teal!13}0.87 & \cellcolor{teal!16}0.84 & \cellcolor{teal!10}0.89 & \cellcolor{teal!16}0.84 & \cellcolor{teal!13}0.87 \\
& SummaC\textsubscript{zs} & \cellcolor{teal!32}0.67 & \cellcolor{teal!20}0.79 & \cellcolor{teal!38}0.62 & \cellcolor{teal!24}0.76 & \cellcolor{teal!35}0.65 & \cellcolor{teal!26}0.74 & \cellcolor{teal!39}0.61 & \cellcolor{teal!27}0.73 & \cellcolor{teal!26}0.74 & \cellcolor{teal!9}0.90 & \cellcolor{teal!25}0.75 & \cellcolor{teal!12}0.88 & \cellcolor{teal!18}0.81 & \cellcolor{teal!14}0.86 & \cellcolor{teal!20}0.79 & \cellcolor{teal!13}0.87 & \cellcolor{teal!15}0.85 & \cellcolor{teal!9}0.90 & \cellcolor{teal!14}0.86 & \cellcolor{teal!10}0.89 \\ \bottomrule

\end{tabular}
}
    \caption{Hallucination risk ratio (HRR) between the original and the pruned model (values less than one are {\sethlcolor{teal!22}\hl{highlighted}}, indicating that the pruned model has a lower hallucination risk than the original model), averaged across all data points over the three prompts for each dataset.  \textbf{Bold} values denote significant differences between the pruned and the original model (paired t-test; $p < 0.05$).}
    \label{tab:hallucination_main}
\end{table*}

\subsection{Summarization 
\label{sec:summarization}}

Table \ref{tab:generation_performance_main} shows summarization performance (ROUGE-1/2/L \& BERTScore) across all datasets.\footnote{We obtain comparable results using 50\% unstructured sparsity, which are omitted for brevity.}
We first observe that the original models perform comparably for BERTScore across most datasets. For example, in Legal Contracts, Llama-2 13B records a BERTScore of 84.75 compared to 84.90 from OPT-IML 30B. We only observe larger performance deviations in the case of RCT, with the original Mistral 7B obtaining the highest BERTScore (88.46) and OPT-IML 30B the lowest (83.12). This suggests that all LLMs generate summaries that are equally semantically similar to the reference summary.
Compared to BERTScore, the scores of the original models in lexical overlap metrics (ROUGE-1/2/L) differ largely not only across models, but also across datasets. For example, Llama-2 7B achieves the second highest ROUGE-L score in RCT (33.50) and the lowest score in FactCC (11.51). Similarly, in RCT, Mistral 7B records an increase of 34.65 (46.16) for ROUGE-L, making it the best performing original model for this metric.

Comparing the performance between original and pruned models, we find that they perform comparably in the majority of cases. For SparseGPT, the summaries score significantly higher (across all metrics) than those from the original model in 19 out of 100 comparisons, while they score significantly lower in 11 out of 100 (\textbf{bold} scores; paired t-test; $p < 0.05$). The results are similar for Wanda, where pruned models perform significantly higher in 20 out of 100 comparisons and significantly lower (\underline{underlined} scores) in 26 out of 100. 
We also find that models pruned with SparseGPT perform more consistently compared to those pruned using Wanda. For example, Llama-2 7B pruned with SparseGPT records a BERTScore of 84.17 for Legal Contracts, compared to 81.52 with Wanda, and 84.73 from the original. 

Comparing across model sizes for Llama-2, pruning seems to be less impactful as model size increases. 
For SparseGPT, we find that the pruned model is comparable (by any metric) in 15 out of 20 comparisons for Llama-2 7B, 18 out of 20 for Llama-2 13B, and in all 20 for Llama-2 70B.

These findings suggest that the summarization performance between pruned and original models is at least comparable.

\subsection{Hallucination Risk}\label{sec:results_hallucination}

Table \ref{tab:hallucination_main} shows the HRR (Section~\ref{sec:hrr}) for all models and datasets, using each hallucination risk metric.\footnote{For reproducibility and transparency, we include the full results (i.e. absolute hallucination risk scores) \texttt{\href{https://docs.google.com/spreadsheets/d/e/2PACX-1vRsxweIz9RfIce2k1Kf9geF1RVyKhBucZZkvp0L0B_-S1QIEQA-mc1zJErHLhf4JWfgyNybv-Ea47f-/pubhtml}{in this link}} due to space constraints.}

\paragraph{Pruning reduces hallucination risk.} 
In almost all cases, irrespective of the pruning method or sparsity pattern (i.e. 2:4 or 50\%), the results show that pruned models have a lower hallucination risk (i.e. {\sethlcolor{teal!22}\hl{values}} lower than 1.0). We find only a single exception, Llama-2 7B pruned with SparseGPT (2:4) for Legal Contracts, with a SummaC\textsubscript{ZS} ratio of 1.01. More importantly, pruned models record significantly lower HRRs (paired t-test; $p < 0.05$). This applies to 284 out of 300 total comparisons across datasets, models, pruning methods, and sparsity patterns.
For example, we observe significantly lower scores across all metrics for Llama-2 7B with SummEval. In particular, SummaC\textsubscript{ZS} scores more than halve for 2:4 semi-structured SparseGPT (0.55) and 2:4 semi-structured Wanda (0.49).

These findings seem counter intuitive, considering that pruned models typically perform comparably to original models in summarization (Table~\ref{tab:generation_performance_main}). 
As both language modeling and summarization performance remains comparable, we hypothesize that \emph{the parametric knowledge removed by pruning \citep{namburi-etal-2023-cost}
``forces'' the model to rely more on the source document during generation and in turn reducing hallucination risk.}
We examine this further in Section \ref{sec:sparsity_impact}.

\paragraph{Semi-structured pruning mitigates hallucination risk.} 

We observe consistently lower HRRs when pruning with semi-structured sparsity (2:4 pattern), versus unstructured pruning at the same sparsity level (50\%). Semi-structured pruning records a lower HRR across all three metrics in 59 out of 65 cases with SparseGPT, and in 55 out of 65 cases with Wanda. We note that semi-structured pruning sometimes produces a substantially lower HRR than unstructured pruning. For example, semi-structured pruning for Llama-2 13B with Wanda records an average SummaC\textsubscript{ZS} HRR of 0.61 versus 0.73 with unstructured pruning.

Unstructured pruning allows weights to be removed in any pattern, enabling pruning according to the optimal layer-wise solution. In contrast, semi-structured pruning constrains the solution space to only the subset that satisfies the desired sparsity pattern (e.g. 2:4, removing two weights in every contiguous block of four). Inevitably, even influential weights with relatively high layer-wise saliency scores may be removed. As semi-structured pruning deviates from the optimal layer-wise solution, a higher proportion of important weights are therefore removed. 
This likely includes relevant parametric knowledge \citep{namburi-etal-2023-cost}, potentially requiring such models to rely more on the source document for generation.

To investigate this, we compute lexical overlap (using ROUGE-1/2/L) between summaries and their source documents across all models, datasets and pruning methods. We find that summaries from models pruned with 2:4 sparsity result in higher lexical overlaps in 114 out of 150 comparisons (three ROUGE metrics, five datasets, five models, two pruning methods) compared to models with 50\% unstructured pruning, supporting our hypothesis.

\paragraph{SummaC and HaRiM\textsuperscript{+} moderately agree.} 
Considering the average results across datasets, we observe mixed signals from SummaC-based HRRs versus HaRiM\textsuperscript{+} HRRs. 
For example, SummaC\textsubscript{Conv} with SparseGPT (2:4) shows that on average, Llama-2 7B benefits most over the original (0.70), followed by Llama-2 13B (0.74). On the contrary, for HaRiM\textsuperscript{+}with 2:4 sparsity, summaries from Llama-2 13B appear to yield the largest reductions in hallucination risk on average (0.81 with SparseGPT and 0.73 with Wanda), followed by OPT-IML 30B (0.86 with both SparseGPT and Wanda). As the results between hallucination risk metrics differ, we want to shed light on how well they agree with each other. Therefore, we compute Pearson's correlation coefficient between all HRR metrics, across all datasets, models and pruning methods. 
Unsurprisingly, both SummaC-based metrics show a strong correlation between them (0.82 averaged across all datasets, models and pruning methods). We also find moderate correlations between HaRiM\textsuperscript{+} and SummaC metrics (0.45 between HaRiM\textsuperscript{+} and SummaC\textsubscript{ZS}; 0.53 between HaRiM\textsuperscript{+}and SummaC\textsubscript{Conv}). 

This is expected, as each metric group computes hallucination risk with different motivations (SummaC-based metrics use entailment methods over the summary and document, while HaRiM\textsuperscript{+} uses token-level predictive likelihood). This explains partly the moderate correlation between them, also \textit{highlighting that it can be beneficial to use HaRiM\textsuperscript{+} and SummaC in conjunction}.

\begin{figure*}
    \centering
    \includegraphics[width=\linewidth]{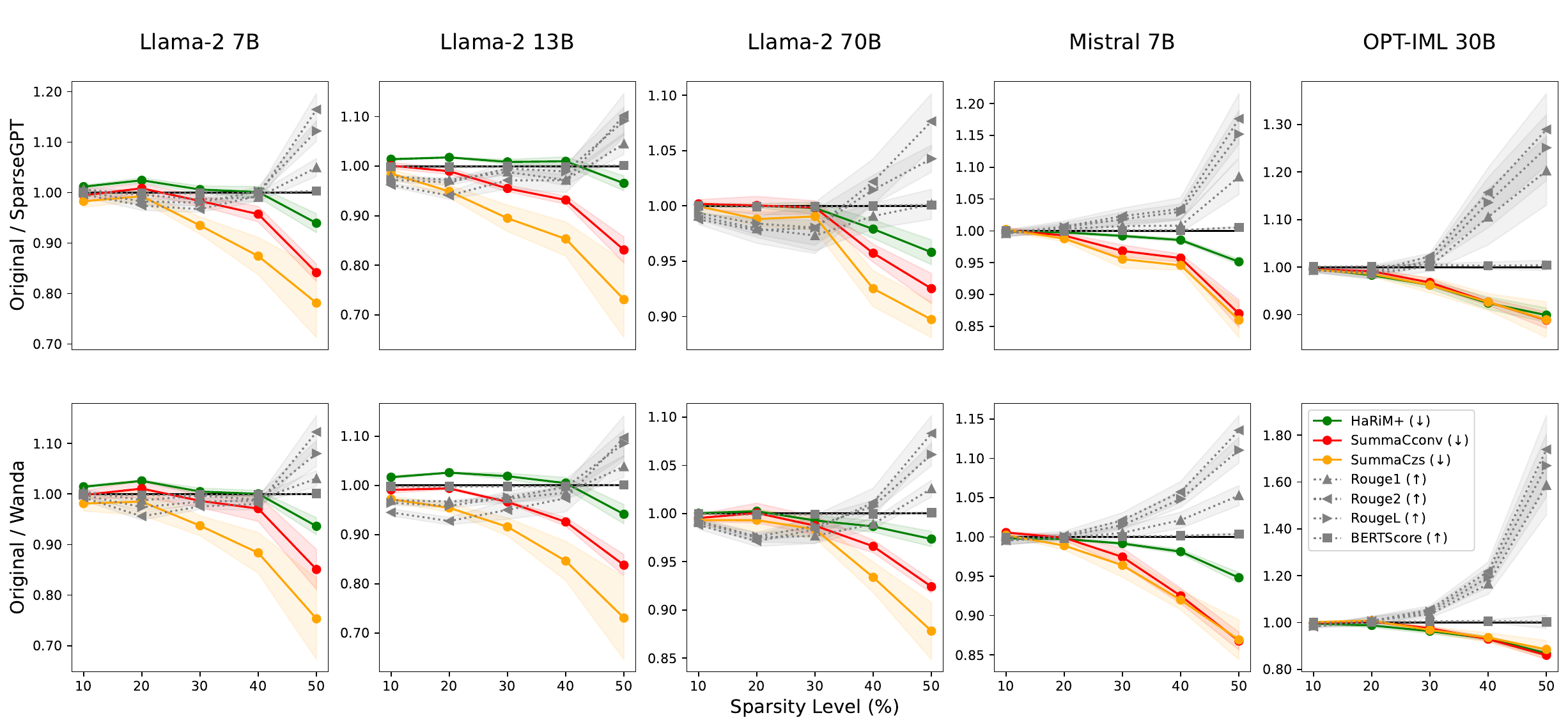}
    \caption{Ratio between a pruned model and the original across five sparsity levels, three hallucination risk metrics (lines with circled markers; lower means pruned is better) and four summary generation performance metrics (gray dotted lines; higher means pruned is better). The ratio for each metric is averaged across all datasets, with error bars indicating standard deviation.}
    \label{fig:across_sparsity_results_overall}
\end{figure*}

\begin{table}[!t]
    \centering
    \footnotesize
        \begin{tabular}{lcccc}
            \toprule
            & Halluc.  & Omiss. & Repet. & Align. \\
            Model & Q1 ($\downarrow$) & Q2 ($\downarrow$) & Q3 ($\downarrow$) & Q4 ($\uparrow$) \\ \midrule
             
            Llama-2 7B & 31 & \textbf{5} & \textbf{0} & \textbf{28} \\ 
            w/ SparseGPT & \textbf{14} & 18 & 9 & 21\\ \cmidrule(lr){2-5}
            \multicolumn{1}{c}{IAA ($\kappa$)} &  0.82 & 0.63 & 0.62 & 0.53 \\ \midrule 
            
            Mistral 7B & 12 & \textbf{9} & \textbf{0}& \textbf{31} \\
            w/ SparseGPT & \textbf{10} & 13 & 5 & 23 \\ \cmidrule(lr){2-5}
            
            \multicolumn{1}{c}{IAA ($\kappa$)} & 0.87 & 0.61 & 0.67 & 0.59\\ \bottomrule
        \end{tabular}
    \caption{Human evaluation results. Values denote the number (out of 100) of summary preferences by participants for the corresponding category. \textbf{Bold} denotes the best performing model per question.}    
    \label{tab:humman_annotation_tasl}
\end{table}

\subsection{Human Evaluation}
\label{ssec:human_eval_results}

Table \ref{tab:humman_annotation_tasl} shows human evaluation results for the questions presented in Section \ref{sec:methodology}.
To offer a fair selection of models, we use summaries generated by the pair that benefited the most (Llama-2 7B) and the least (Mistral 7B) in terms of hallucination risk (i.e. the largest and smallest improvements in Table \ref{tab:hallucination_main}). We then select the corresponding summaries from the pruned counterpart, specifically SparseGPT (2:4) which obtained the most consistent summarization performance (Section \ref{sec:summarization}).

\paragraph{Original models hallucinate more.} 
Summaries generated by the original Llama-2 7B model contain hallucinations in 31 cases (out of 100) compared to 14 with SparseGPT applied. 
In comparison, the results for Mistral 7B also suggest that 10 (out of 100) summaries from Mistral 7B pruned with SparseGPT contain hallucinations, compared to 12 summaries generated using the original model (i.e. a smaller difference compared to Llama-2 7B). 

This aligns well with our initial expectations and HRR results (Table \ref{tab:hallucination_main}), as Mistral 7B benefits less from pruning in terms of hallucination risk compared to Llama-2 7B. For example, considering SummaC\textsubscript{ZS} for SummEval, Llama-2 7B pruned with SparseGPT approximately halves the hallucination risk (0.49) compared to 0.79 with Mistral 7B. From analyzing human evaluation results, we found that the large difference between pruned and original Llama-2 7B is predominantly driven by major factual errors (discussed in Section \ref{sec:qualitative_analysis}).

\paragraph{Original models omit and repeat slightly less.} With substantial (0.61-0.80) agreement between participants, the results agree that both original models had no repetitions in their summaries and omitted less important information compared to pruned model summaries (e.g. nine instances with Mistral 7B compared to 13 with its pruned version with SparseGPT).

Comparing how well the summaries semantically align with the source document, the results show a preference towards the original models (with moderate agreement; 0.40-0.60). For example, 28 (out of 100) summaries of the original Llama-2 7B were selected as more aligned compared to 21 summaries when pruned with SparseGPT.

\section{Impact of Pruning Sparsity on Hallucination Risk}\label{sec:sparsity_impact}

To better understand previous observations and test our hypothesis (i.e. sparsity likely encourages models to focus more on the source document during generation), we analyze hallucination risk across different sparsity levels. We additionally track the lexical overlap (using ROUGE-1/2/L) and semantic overlap (using BERTScore) between the generated summary and the source document. Our hypothesis is: \textit{if lexical overlap positively correlates with sparsity levels, it suggests that pruned models may rely more on the source document for generation.}

Figure \ref{fig:across_sparsity_results_overall} shows the summarization performance ratio (ROUGE-1/2/L and BERTScore; ratio computed as pruned over original) and HRR ($\downarrow$) for five LLMs and two pruning methods, across increasing levels of unstructured sparsity (10\% to 50\%). We only consider unstructured sparsity, since the 2:4 semi-structured pattern enforces a fixed sparsity level of 50\%. The ratio for each metric is averaged across datasets for brevity, with error bars indicating standard deviation. For summarization performance, a ratio higher than 1.0 indicate that the pruned model performs better than the original, whereas a HRR lower than 1.0 indicates that summaries from the pruned model have a lower hallucination risk.

\paragraph{Hallucination risk reduces as sparsity increases.} Results consistently show that hallucination risk reduces as sparsity levels increase, across all models and pruning methods. For example, with Llama-2 13B and Wanda, SummaC\textsubscript{ZS} HRR reduces from 0.98 at 10\% sparsity, to 0.90 at 30\% to finally 0.73 at 50\%. Moreover, OPT-IML 30B displays a remarkably linear improvement (i.e. with SparseGPT the HRR is 1.00 at 10\% sparsity, 0.95 at 30\% and 0.90 at 50\%, for all hallucination risk metrics). These findings suggest that \emph{increasing sparsity to moderate levels (up to 50\%) does indeed appear to reduce hallucination risk in generated summaries}.

\paragraph{Semantic and lexical overlaps differ.} Observing the lexical (ROUGE) and semantic (BERTScore) similarity ratios between document and generated summary across sparsity levels, the outcomes are mixed. In almost all cases for both pruning methods, BERTScore results remain comparable to the original model (close to 1.0) up to 50\% sparsity, with minimal deviation across datasets. This shows that summaries from pruned models are as semantically similar to the source document as those from original models, across all sparsity levels.

However, there is a stark contrast with ROUGE-1/2/L. For Llama-2 models, ROUGE-based ratios appear to decrease until 30\% sparsity, then increase substantially and peak above 1.0 (the original model baseline) at 50\% sparsity. For Mistral 7B and OPT-IML 30B, we observe that ROUGE-based ratios increase above 1.0 (higher than original) from a lower sparsity (20\%). As summaries from pruned models remain as semantically similar to the source document as those from original models, their \emph{higher lexical overlap with the source document indicates that pruned models focus more on the input document to generate a summary}.

\begin{table}[t]
    \scriptsize
    \centering
        \begin{tabular}{lcc}
        \toprule
        & \multicolumn{2}{c}{ROUGE-1/2/L} \\ \cmidrule(lr){2-3}
        Model & SparseGPT & Wanda \\ \midrule
         Llama-2 7B &  -0.69 / \textbf{-0.89} / \textbf{-0.90} & -0.45 / \textbf{-0.86} / -0.79 \\
         Llama-2 13B & -0.70 / -0.77 / -0.84 & -0.72 / -0.78 / \textbf{-0.85} \\
         Llama-2 70B & -0.39 / \textbf{-0.86} / \textbf{-0.84} & -0.69 / \textbf{-0.86} / \textbf{-0.86}\\
         Mistral 7B & \textbf{-0.91} / \textbf{-0.97} / \textbf{-0.97} & -0.88 / \textbf{-0.96} / \textbf{-0.97} \\
         OPT-IML 30B & -0.70 / \textbf{-0.93} / \textbf{-0.89} & \textbf{-0.93} / \textbf{-0.94} / \textbf{-0.93} \\
         \bottomrule
        \end{tabular}
    \caption{Averaged Pearson's correlation coefficient ($r$) between hallucination risk and ROUGE-based metrics (calculated between the generated summaries and the source documents) across sparsity levels. \textbf{Bold} values indicate significant correlations ($p < 0.05$).}
    \label{tab:correlation_results}
\end{table}

\paragraph{Higher lexical overlap, lower hallucination risk.} Surprisingly, we observe an inversely proportional relationship between ROUGE-based ratios and HRRs. 
We hypothesize that a higher lexical overlap with the source document is a possible reason for the lower hallucination risk. 
To assess this, we calculate Pearson's correlation coefficient, averaged across sparsity levels between all HRR and ROUGE-based metrics (Table \ref{tab:correlation_results}, significant correlations in \textbf{bold}; $p < 0.05$).

We note a strong significant inverse correlation (Pearson's $r < -0.8$) for both pruning methods for ROUGE-2/L across almost all models (excluding Llama-2 13B) and $r < -0.4$ for ROUGE-1. This suggests that \emph{a higher lexical overlap could be responsible for the reduced hallucination risk, while increasing sparsity appears responsible for an increasing lexical overlap}. In particular, we find an almost perfect negative relationship between ROUGE-based ratios and HRRs (-0.97 with SparseGPT) for Mistral 7B.
This corroborates findings from the study by \citet{durmus-etal-2020-feqa}, which shows that summaries with a higher lexical similarity to the source document are less likely to contain hallucinations.

\begin{table*}[!t]
\scriptsize
\centering
\begin{tabular}{p{0.46\linewidth} p{0.46\linewidth}}
\toprule
\multicolumn{1}{c}{\footnotesize Source Document} & \multicolumn{1}{c}{\footnotesize Generated Summary} \\
\midrule

\multicolumn{1}{c}{\textit{FactCC \#205}} & \multicolumn{1}{c}{\textit{Hallucination (Q1)}} \\
{What do we have for the contestant on ``The Price Is Right''? A brand-new car! Whoops. [\textellipsis] \hlpos{model Manuela Arbelaez} accidentally revealed the correct answer [\textellipsis]} &
{\hlneg{The contestant} on ``The Price Is Right'' accidentally revealed the correct answer to a guessing game for a new Hyundai Sonata [\textellipsis]} \\
\midrule
\multicolumn{1}{c}{\textit{SummEval \#86}} & \multicolumn{1}{c}{\textit{Omission (Q2)}} \\
{[\textellipsis] Patrick Kluivert has made a winning start in \hlpos{World Cup} qualifying as coach [\textellipsis] The Caribbean island team advanced to the second qualifying round in CONCACAF for the \hlpos{2018 tournament} [\textellipsis]} &
{Patrick Kluivert [\textellipsis] has successfully led Curacao to the second qualifying round in CONCACAF for \hlneg{the 2018 tournament}. [\textellipsis]} \\
\midrule
\multicolumn{1}{c}{\textit{SummEval \#33}} & \multicolumn{1}{c}{\textit{Repetition (Q3)}} \\
{[\textellipsis] auction at Sotheby's on Wednesday. While \hlpos{the 13 letters were expected to bring in anywhere from \$300,000 to \$600,000}, [\textellipsis] \hlpos{estimated to bring in anywhere from \$300,000 to \$600,000}, but [\textellipsis]} & 
{ [\textellipsis] \hlneg{The 13 letters were expected to bring in anywhere from \$300,000 to \$600,000, but received no bids}. [\textellipsis]  \hlneg{The letters were estimated to be worth between \$300,000 and \$600,000}. [\textellipsis] }  \\
\midrule
\multicolumn{1}{c}{\textit{FactCC \#136}} & \multicolumn{1}{c}{\textit{Alignment (Q4)}} \\
{Deion Sanders is such a dad. \hlpos{The NFL legend called out Deion Sanders Jr. on Twitter for saying he only eats ``hood doughnuts.''} In response, the elder Sanders [\textellipsis] reminded his son he has a trust fund, a condo and his own clothing line called ``Well Off'' [\textellipsis]} &
{\hlneg{Deion Sanders Jr. is a Southern Methodist University wide receiver and aspiring entrepreneur who loves ``hood doughnuts''} and has a trust fund, condo, and clothing line called ``Well Off.'' [\textellipsis]} \\
\bottomrule
\end{tabular}
\caption{Examples of \hlneg{errors} in generated summaries corresponding to the human evaluation questions and the \hlpos{context} from the source document.}
\label{tab:qualitative_examples}
\end{table*}

\section{Qualitative Analysis\label{sec:qualitative_analysis}}

Following the human evaluation (see Sections \ref{sec:human_evaluation_method} and \ref{ssec:human_eval_results}), we review specific cases, highlighting issues with the summaries generated by pruned models in Table \ref{tab:qualitative_examples}.

\paragraph{Hallucinations.}
Our analysis of the human evaluation task results suggests that hallucinations in the summaries from both Llama-2 7B and Mistral 7B are either: (a) additional information not supported by the source document, or (b) modified or misplaced information from the source document (e.g. FactCC \#205).

\paragraph{Omissions.}
Omission is a category where we found a few instances of disagreement between the participants. In general, participants agree in clear cases like SummEval \#86 (e.g. \textit{``2018 tournament''} should be \textit{``2018 World Cup''}). Comparatively in disagreements, omitted information is more nuanced and difficult to detect, such as important details from the source document (e.g. missing dates).

\paragraph{Repetitions.}
Interestingly, we find that summaries containing repetitions occur when the source document also contains repeating information (e.g. the price range ``\textit{\$300,000 to \$600,000}'' duplicated in SummEval \#33).

\paragraph{Alignment.}
The generated summaries that are less aligned to the source document do not necessarily contain any hallucinations, omissions, or repetitions. However, we found that they do not entirely convey the original meaning of the source document. For example in FactCC \#136, the source describes \textit{Deion Sanders Jr.} being publicly scolded by his father for downplaying his wealthy lifestyle.
However, this particular piece of information is not conveyed in the generated summary.

\section{Conclusion}
We conducted an extensive study to assess the hallucination risk of LLMs after pruning. We experimented with two state-of-the-art pruning methods applied to five instruction-tuned LLMs. We measured the hallucination risk using three established automatic metrics, in addition to a human evaluation. Our results show that as models are pruned to moderately high sparsity levels, the risk of generating hallucinating content decreases. Our analysis suggests that pruned models tend to generate summaries that have a greater lexical overlap with the source document, offering a possible explanation for the lower hallucination risk. 

In future work, we plan to explore the relationship between hallucination risk and model quantization \citep{dettmers-etal-2022-gpt3,frantar-etal-2023-optq} and also expand to tasks such as open-book question answering \citep{ciosici-etal-2021-perhaps} and machine translation \citep{guzman-etal-2019-flores}. Finally, an interesting direction is to investigate the relationship between hallucination risk and explanation faithfulness~\citep{chrysostomou-aletras-2022-empirical,zhao-aletras-2023-incorporating}.

\section*{Acknowledgments}
We would like to thank the anonymous reviewers and action editor for their invaluable feedback. 
MW is supported by the Centre for Doctoral Training in Speech and Language Technologies (SLT) and their Applications funded by UK Research and Innovation grant EP/S023062/1. ZZ and NA are supported by EPSRC grant EP/V055712/1, part of the European Commission CHIST-ERA programme, call 2019 XAI: Explainable Machine Learning-based Artificial Intelligence. NA is also supported by EPSRC grant EP/Y009800/1, part of the RAI UK Keystone projects.

\bibliography{anthology,custom}
\bibliographystyle{acl_natbib}

\end{document}